\documentclass{article} 

\usepackage[ruled, vlined, linesnumbered]{algorithm2e}
\usepackage{iclr2021_conference,times}

\usepackage{amsmath,amsfonts,bm}

\def\eqref#1{equation~\ref{#1}}

\def\1{\bm{1}}

\DeclareMathAlphabet{\mathsfit}{\encodingdefault}{\sfdefault}{m}{sl}
\SetMathAlphabet{\mathsfit}{bold}{\encodingdefault}{\sfdefault}{bx}{n}

\usepackage{hyperref}
\usepackage{tabularx}
\usepackage{url}
\usepackage{amssymb}
\usepackage{courier}

\usepackage{graphicx}
\usepackage{pgf, tikz}
\usetikzlibrary{arrows, shapes, automata}

\DeclareMathOperator*{\argminB}{argmin}

\title{Active WeaSuL: Improving Weak Supervision with Active Learning}

\author{Samantha Biegel$^{1,2}$, Rafah El-Khatib$^{1,3}$, Luiz Otavio Vilas Boas Oliveira$^{1}$, Max Baak$^{1}$ \& Nanne Aben$^{1}$ \\
$^1$ ING Wholesale Banking Advanced Analytics, Amsterdam, The Netherlands \\
$^2$ University of Amsterdam, Amsterdam, The Netherlands \\
$^3$ Feebris, London, United Kingdom \\
\texttt{samantha.r.biegel@gmail.com} \\
\texttt{rafah@feebris.com} \\
\texttt{\{luiz.vilas.boas.oliveira, max.baak, nanne.aben\}@ing.com}
}

\iclrfinalcopy 
\begin{document}

\maketitle

\begin{abstract}
The availability of labelled data is one of the main limitations in machine learning. We can alleviate this using weak supervision: a framework that uses expert-defined rules $\boldsymbol{\lambda}$ to estimate probabilistic labels $p(y|\boldsymbol{\lambda})$ for the entire data set. These rules, however, are dependent on what experts know about the problem, and hence may be inaccurate or may fail to capture important parts of the problem-space. To mitigate this, we propose Active WeaSuL: an approach that incorporates active learning into weak supervision. In Active WeaSuL, experts do not only define rules, but they also iteratively provide the true label for a small set of points where the weak supervision model is most likely to be mistaken, which are then used to better estimate the probabilistic labels. In this way, the weak labels provide a warm start, which active learning then improves upon. We make two contributions: 1) a modification of the weak supervision loss function, such that the expert-labelled data inform and improve the combination of weak labels; and 2) the maxKL divergence sampling strategy, which determines for which data points expert labelling is most beneficial. Our experiments show that when the budget for labelling data is limited (e.g. $\leq 60$ data points), Active WeaSuL outperforms weak supervision, active learning, and competing strategies, with only a handful of labelled data points. This makes Active WeaSuL ideal for situations where obtaining labelled data is difficult.\footnote{Code for Active WeaSuL: \url{https://github.com/SamanthaBiegel/ActiveWeaSuL}}

\end{abstract}

\section{Introduction}
\pagenumbering{arabic}
Machine learning models often require large amounts of labelled data to work properly. However, collecting large amounts of high-quality labelled data is not always straightforward. While in some cases we may be able to outsource the labelling process at a competitive price, we often require the use of expensive domain experts to do the labelling, for example due to the inherent difficulty of the task or due to privacy concerns preventing us from sharing the data externally. In these cases, it is imperative that we explore ways in which we can make use of our domain experts in a more efficient way.

One way to do so is through a process called weakly supervised learning (or weak supervision for short) \citep{Zhou2018}. In this process, we ask domain experts to define labelling functions: rules that they think are indicative of a given class, such as ``IF $X_1 < 5$ THEN $y = 1$'' (Figure 1A). These labelling functions are applied to the data, resulting in a set of weak labels $\boldsymbol{\lambda}$, which are then combined using a generative model, resulting in the probabilistic labels $p(y | \boldsymbol{\lambda})$ for all data points (Figure 1B). Finally, the probabilistic labels are used as a proxy to train a discriminative model that predicts $\textbf{y}$ from the feature matrix $\mathbf{X}$ (Figure 1D). In short, weak supervision allows one to train a supervised model using expert-defined rules.

While weak supervision has shown to be very powerful \citep{ratner2016data, Bach2019, Badene2019, Dunnmon2020}, we observe that it does not always reach its optimal performance. For example, Figure 1D shows that the predicted decision boundary is rotated with respect to the optimal decision boundary. In practice, this may happen when the expert-defined rules are imprecise or cover only parts of the problem-space, when the conditional independence between weak labels is not properly specified, or due to biases introduced by the generative model.

\begin{figure}[h!]
    \centering
    \includegraphics[width=\textwidth]{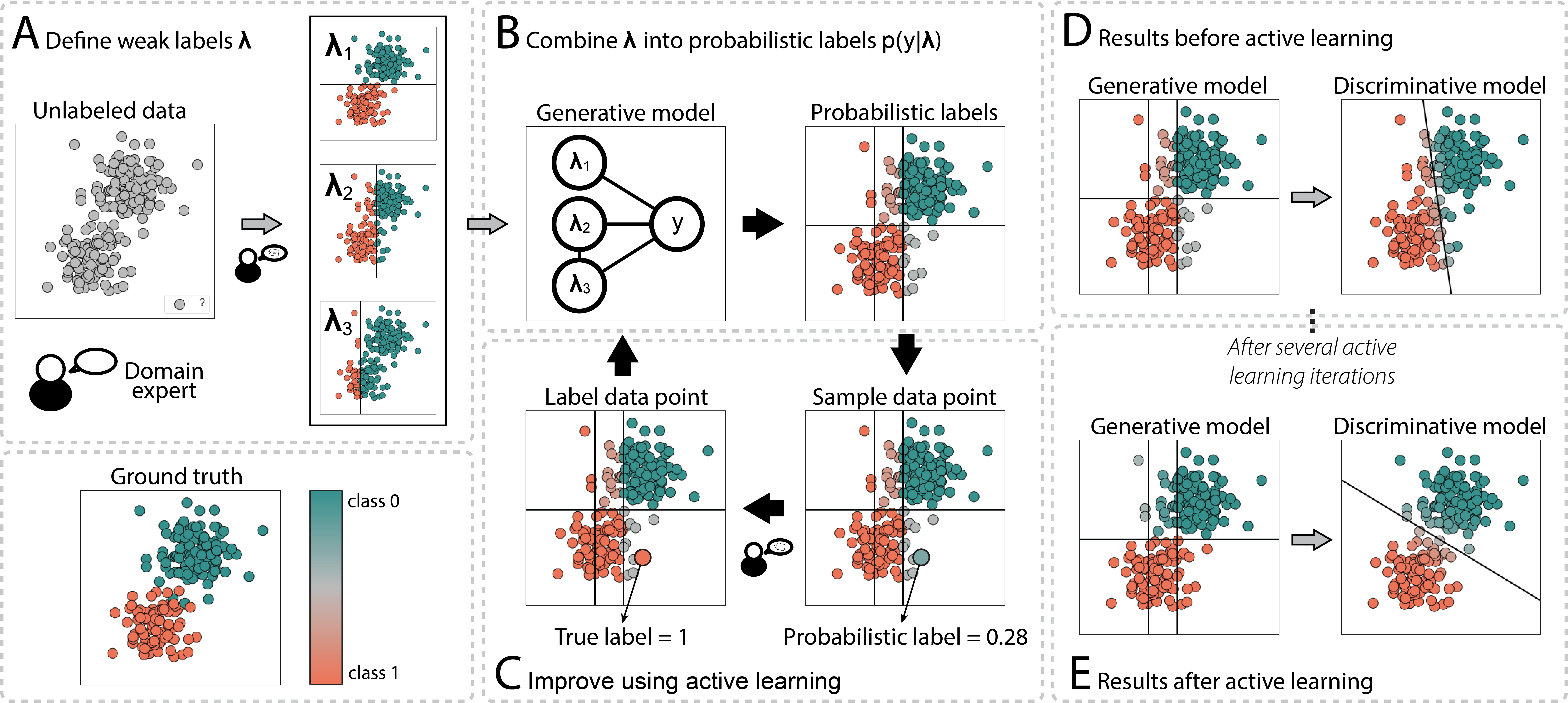}
    \caption{Overview of the Active WeaSuL approach. A) A domain expert defines the labelling functions, which when applied to data are converted to weak labels $\boldsymbol{\lambda}$. B) A generative model is used to combine the weak labels $\boldsymbol{\lambda}$ into probabilistic labels $p(y | \boldsymbol{\lambda})$. C) We use an active learning approach to iteratively select the best data point for the expert to label. This information is then used to improve the estimation of $p(y | \boldsymbol{\lambda})$. D\&E) Predictions $p(y | \boldsymbol{X})$ of a discriminative model trained to predict probabilistic labels $p(y | \boldsymbol{\lambda})$ from the feature matrix $\textbf{X}$ both D) before active learning and E) after active learning.}
    \label{fig:active-weasul}
\end{figure}

Inspired by \citet{nashaat2019hybridization}, we propose to address this problem by incorporating a second kind of domain expert knowledge: besides the expert-defined rules, we ask domain experts to label a small number of data points (Figure 1C). Specifically, we do this in an active learning fashion, in which we iterate over: 1) determining for which data point the model would benefit most from knowing the true label $y$; 2) asking an expert to label this data point for us, which we consider as the ground truth label; and 3) using this labelled data point to improve the estimation of the probabilistic labels $p(y | \boldsymbol{\lambda})$. Optionally, one can also do a fourth step, which is to train a discriminative model on the obtained probabilistic labels, using the data's features. In this way, over the course of several iterations, we gradually improve the performance of both the generative and discriminative models (Figure 1E).

We call our approach Active WeaSuL, as a nod to the combination of active learning (i.e. iteratively learning from a small number of expert-labelled data points) and weakly supervised learning (i.e. learning from expert-defined rules). To combine these two methodologies, we make two important contributions: 1) a modification to the weak supervision loss function, allowing us to use expert-labelled data to improve the combination of weak labels; and 2) the maxKL divergence sampling method, a query strategy to determine for which data points expert-labelling would be most beneficial within this framework.

We benchmark Active WeaSuL against three alternative sets of models on three data sets. We first show that Active WeaSuL outperforms the three baseline approaches; i.e. active learning by itself, weak supervision by itself, as well as the original approach to combining active learning and weak supervision by \citet{nashaat2019hybridization}, in particular when we only have budget to label a limited number of data points. Subsequently, we show the benefits of the maxKL divergence sampling strategy compared to either random sampling or sampling based on the margin from the decision boundary.

\section{Related Work}
The combination of active learning and weak supervision has been pioneered by \citet{nashaat2019hybridization}. In their work, they also iteratively obtain expert-labelled data which they then use to improve the weak supervision model. The most important difference with our work lies in the way in which the expert-labelled data is incorporated into the weak supervision model. While Active WeaSuL uses the expert-labelled data to learn how to best combine the weak labels (thereby affecting predictions for all data points), \citet{nashaat2019hybridization} instead use the expert-labelled data to correct the weak labels for the corresponding individual data points. Specifically, if we have three labelling functions and for one data point $i$ we have the weak labels $\boldsymbol{\lambda}_{i}=[1,1,0]$ and an expert-label $\textbf{y}_i=1$, then they correct the weak labels with the expert-label by setting $\boldsymbol{\lambda}_{i}=[1,1,1]$, whereas Active WeaSuL leaves the weak labels unchanged and instead improves the way in which the weak labelling functions are combined.

We note that \citet{nashaat2019hybridization} have also released two follow-up works \citep{nashaat2020asterisk, nashaat2020wesal}. In these works, they use the same combination of active learning and weak supervision as in their original work. In \citet{nashaat2020asterisk}, they additionally use a mechanism to automatically generate labelling functions based on a small set of labelled data. In this work, we assume that we have no labelled data at the beginning, and that the labelling functions are defined by experts, hence we compare Active WeaSuL to the original work by \citet{nashaat2019hybridization}.

Besides the approach outlined above, there are several others that combine active learning and weak supervision, though with a different goal in mind. For instance, several authors have proposed to use labelled data to improve how the labelling functions are defined \citep{boecking2020interactive, awasthi2020learning, chen2019scene, cohen2019interactive, varma2018snuba}. A nice example of this is the work of \citet{boecking2020interactive}, who propose a method that automatically improves the set of labelling functions based on feedback from users. In this work, we take a different angle: we assume the labelling functions themselves are fixed and instead use the expert-labelled data to improve how we combine the weak labels.

Alternatively, several authors have proposed to use weak supervision to improve active learning \citep{Brust2020, Gonsior2020}, rather than the other way around. In this approach, for a small set of data points, each point is labelled by several experts. These labels are then combined into one estimate per data point using weak supervision (i.e. by considering each expert as a labelling function). Such an approach works well in scenarios where experts cannot reliably derive the true label. We note that this serves a different goal than our work: here, we assume that the experts are in fact able to reliably determine the true labels, but that the budget for labelling is limited. We instead focus on using weak supervision to incorporate expert-defined rules into our model, thereby gaining a warm start over active learning by itself, which otherwise needs to start from scratch.

\section{Methods}\label{sec:methods}
 \subsection{Weak supervision}
In weak supervision, we estimate probabilistic labels $p(y | \boldsymbol{\lambda})$ from a set of weak labels $\boldsymbol{\lambda}$ using a generative model. Here, we use the generative model as formulated by \citet{ratner2019training}. Let $\boldsymbol{\lambda}$ be the $n \times m$ matrix of weak labels, where $n$ is the number of samples and $m$ is the number of weak labels; let $\boldsymbol{\lambda}_{i,*}$ be the $i$th row of this matrix (i.e. all weak labels for data point $i$); and let $\boldsymbol{\lambda}_{*,j}$ be the $j$th column of this matrix (i.e. all data points for weak label $j$). Consider $\boldsymbol{\Sigma}$ the $m \times m$ covariance matrix of $\boldsymbol{\lambda}$. Consider $\textbf{y}$ the $n \times 1$ vector of binary ground truth labels.

Next, let us define two assumptions. First, let us assume that we know the class prior $p(y)$. We can obtain $p(y)$ either through expert knowledge or by estimating it from the data \citep{ratner2019training}. Second, we assume to know which weak labels are conditionally independent of each other given $y$. We denote this with the set $\Omega$, where weak label $i$ and weak label $j$ are conditionally independent given $y$ for all $(i,j) \in \Omega$. We can either obtain $\Omega$ through expert knowledge, for example because we know which labelling functions are conceptually unrelated, or we can estimate $\Omega$ from the data \citep{varma2019learning}.

Given the conditional independence structure $\Omega$, we can solve the matrix completion problem by solving the objective problem from \citet{ratner2019training} as:
\begin{gather}
    \label{eq:z-min}
    \hat{\textbf{z}} = \argminB_{\textbf{z}} ||(\boldsymbol{\Sigma}^{-1} + \textbf{zz}^T)_{\Omega}||_F
\end{gather}


where $(.)_\Omega$ is equivalent to $(.)_{(i,j)}$ for $(i,j)$ in $\Omega$ and $||.||_F$ is the Frobenius norm. Given the class prior $p(y)$ and the conditional independence structure $\Omega$, we can then define a function $f$ that transforms the parameters $\hat{\textbf{z}}$ into probabilistic labels as:
\begin{align}
\hat{p}(y | \boldsymbol{\lambda}_{i,*}) = f(\hat{\textbf{z}}, \boldsymbol{\lambda}_{i,*})
\end{align}

For more details on these equations, we refer to \citet{ratner2019training}.

\subsection{Modifying the generative model loss function to incorporate active learning}
In this work, we use active learning to improve the generative model's estimate of $p(y | \boldsymbol{\lambda}_{i,*})$. To this end, we modify the objective function to include a penalty $Pe(\boldsymbol{z})$ that nudges the optimal parameters towards a configuration where $p(y | \boldsymbol{\lambda}_{i,*}) = \textbf{y}_i$ for all expert-labelled data points $i$.

\begin{align}
    \label{eq:z-min-penalty}
    \hat{\textbf{z}} = \argminB_{\textbf{z}} ||(\boldsymbol{\Sigma}^{-1} + \textbf{zz}^T)_{\Omega}||_F + \alpha Pe(\textbf{z})
\end{align}


The hyper-parameter $\alpha$ can be tuned such that the penalty is competitive in size with the original loss term. 
The penalty $Pe(\textbf{z})$ is defined as the quadratic difference between the probabilistic label $p(y | \boldsymbol{\lambda}_{i,*}) = f(\textbf{z}, \boldsymbol{\lambda}_{i,*})$ and the true (binary) label $\textbf{y}_i$ for all data points $i$ in the set of expert-labelled data $D$ obtained through active learning:
\begin{align}
Pe(\textbf{z}) = \sum_{i \in D} (f(\textbf{z}, \boldsymbol{\lambda}_{i,*}) - \textbf{y}_i)^2
\end{align}

In short, the penalty encourages the generative model to combine the weak labels $\boldsymbol{\lambda}$ such that the resulting probabilistic labels are concordant with the expert-labelled data, i.e. $p(y | \boldsymbol{\lambda}_{i,*}) = \textbf{y}_i$ for all expert-labelled data points $i$.

\subsection{The maxKL divergence active learning sampling strategy}
While the modified loss function allows us to improve the estimate of $p(y | \boldsymbol{\lambda}_{i,*})$, it does not tell us which data points we should ask the expert to label. To this end, we define the maxKL divergence active learning sampling strategy, based on Kullback-Leibler (KL) divergence, which we specifically tailor to the combination of active learning and weak supervision. The main idea is that, for each iteration, we sample a data point from the region where the generative model and the expert-labelled data disagree most, as we argue that these data points will be most informative to the model (Figure 2).

\begin{figure}[h!]
    \centering
    \includegraphics[width=0.7\textwidth]{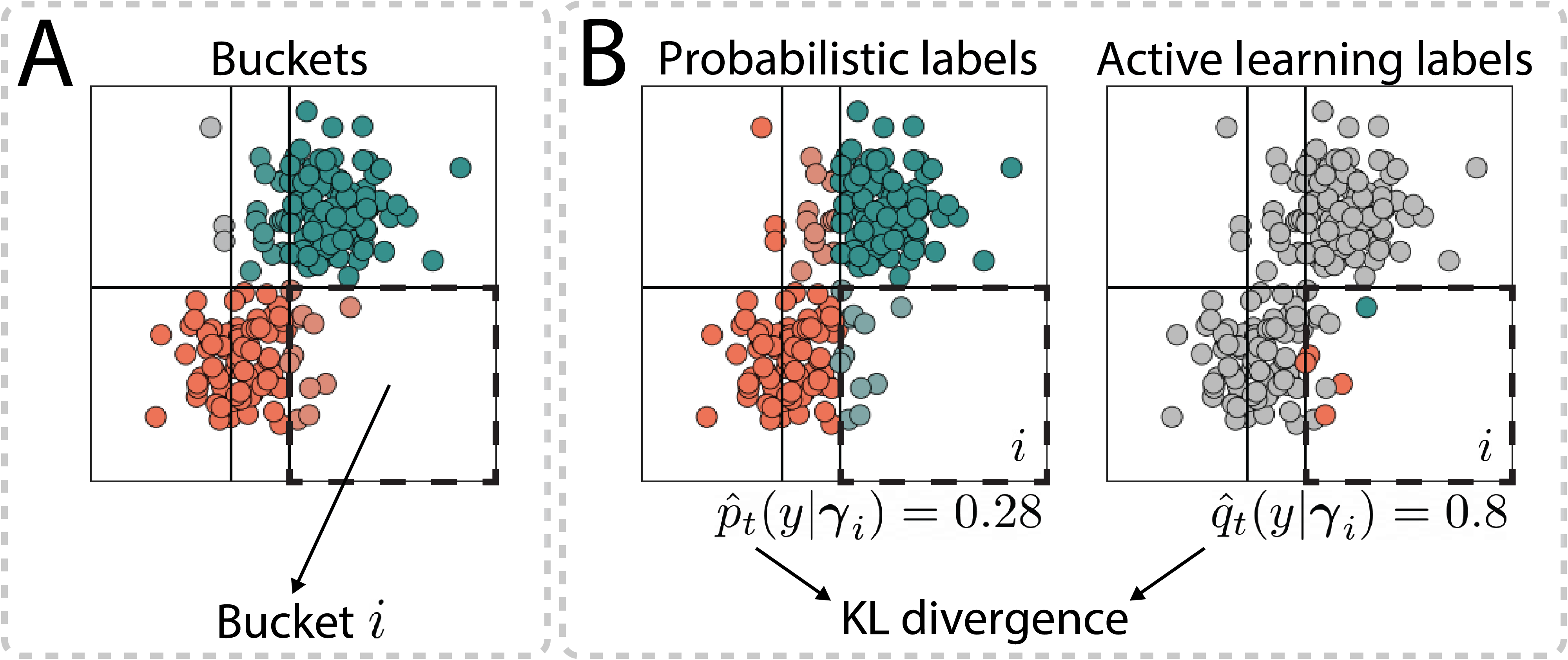}
    \caption{Overview of the active learning sampling method. A) The definition of a \textit{bucket}, i.e. a unique configuration of weak label values. B) For each bucket $i$, we determine the KL divergence between the distributions of the probabilistic labels and the expert-labelled data. We then choose a data point from the bucket with the maximal divergence for expert-labelling in the next active learning iteration.}
    \label{fig:active-weasul}
\end{figure}

Consider $\boldsymbol{\gamma}$ as the $r \times m$ matrix of unique row values in $\boldsymbol{\lambda}$, where each row $\boldsymbol{\gamma}_{i,*}$ is a unique configuration of weak label values. We refer to these rows as \textit{buckets} that split up the problem space (Figure 2A). Let $t$ denote the active learning iteration index. Since we label one data point per active learning iteration, $t$ also denotes the number of data points for which we have collected expert labels so far. Using these concepts, we can define the generative model's estimate of the probability of observing $y$ in a given bucket $i$ at a given iteration $t$ as:
\begin{align}
\hat{p}_t(y | \boldsymbol{\gamma}_{i,*}) = f(\hat{\textbf{z}}, \boldsymbol{\gamma}_{i,*}) \label{eq:prob-labels-t}
\end{align}

where $\hat{\textbf{z}}$ is updated using Equation \ref{eq:z-min-penalty} at every iteration $t$. Additionally, we define a second estimate of the probability of observing $y$ in a given bucket $i$ at a given iteration $t$, this time using expert-labelled data only:
\begin{align}
\hat{q}_t(y | \boldsymbol{\gamma}_{i,*}) = \frac{\sum_{j \in D_i}{\textbf{y}_j}}{|D_i|}
\end{align}

where $D_i$ is the set of expert-labelled data points in bucket $i$. To prevent division by zero, we set $\hat{q}_t(y | \boldsymbol{\gamma}_{i,*}) = \text{round}(\hat{p}_{t=0}(y | \boldsymbol{\gamma}_{i,*}))$, i.e. to the generative model's estimate before any active learning iterations, rounded to the nearest binary value, when $|D_i| = 0$.

When for a bucket $i$ the estimates $\hat{p}_t(y | \boldsymbol{\gamma}_{i,*})$ and $\hat{q}_t(y | \boldsymbol{\gamma}_{i,*})$ are very different from each other, it indicates that the generative model and the expert-labelled data are in disagreement with each other for this particular bucket. This suggests that for the next active learning iteration $t+1$ it would be highly informative to select a data point from that bucket for expert-labelling.

We quantify the difference between $\hat{p}_t(y | \boldsymbol{\gamma}_{i,*})$ and $\hat{q}_t(y | \boldsymbol{\gamma}_{i,*})$ using the KL divergence:
\begin{align}
KL_{t, i}(p||q) &= p \cdot \log \frac{p}{q} + (1-p) \cdot \log \frac{1-p}{1-q} \label{eq:kl-div}
\end{align}

where we use $p = \hat{p}_t(y | \boldsymbol{\gamma}_{i,*})$ and $q = \hat{q}_t(y | \boldsymbol{\gamma}_{i,*})$. To prevent division by zero, we cap $q$ to be within $[\epsilon, 1-\epsilon]$, where $\epsilon$ is a very small number.

We can now define our the maxKL divergence sampling strategy. At iteration $t$, we sample a data point from the bucket $i$ that has the maximum KL divergence $KL_{t, i}$. We then ask an expert to label this data point, resulting in the updated estimates $\hat{p}_{t+1}(y | \boldsymbol{\gamma}_{i,*})$ and $\hat{q}_{t+1}(y | \boldsymbol{\gamma}_{i,*})$ for all buckets $i$. The complete algorithm for Active WeaSul is shown in Supplementary Algorithm 1.

\section{Results}

\subsection{Benchmarking Active WeaSuL on artificial data}

We first benchmark Active WeaSuL on the simple artificial data set illustrated in Figure 1. Essentially, this data set consists of two balanced classes, where each class is modelled by a 2-dimensional Gaussian, with a training set of 10,000 data points and a test set of 3,000 data points. Although the labels are known to us, we do not make them known to the model outside of the active learning steps. We define three labelling functions on these data: $\boldsymbol{\lambda}_2$ and $\boldsymbol{\lambda}_3$ on the first dimension $\textbf{X}_{1}$, and $\boldsymbol{\lambda}_1$ on the second dimension $\textbf{X}_{2}$. Note that $\boldsymbol{\lambda}_2$ and $\boldsymbol{\lambda}_3$ are not conditionally independent since they are both defined on $\textbf{X}_{1}$, and we incorporate this information in the generative model via $\Omega$ (Equation \ref{eq:z-min}). We set $\alpha=1$ in Equation \ref{eq:z-min-penalty} since the loss of the initial generative model is close to zero.

To compare Active WeaSuL with weak supervision by itself, we consider the predictive performance of Active WeaSuL on these artificial data during the first 30 active learning iterations (Figure 3A\&B). To this end, we obtain the predicted labels $\hat{\textbf{y}}$ of the generative model of Active WeaSuL by rounding the probabilistic labels $\hat{p}_t(y | \boldsymbol{\lambda})$ to the nearest binary value, which we then compare to the ground truth labels $\textbf{y}$ to determine the accuracy. For the generative model, we observe that the accuracy improves from 0.81 at $t=0$ (i.e. weak supervision, without any active learning) to 0.96 within 4 active learning iterations (Figure 3A). We observe similar improvements using the discriminative model (Figure 3B) (Supplementary Materials). This shows that combining weak supervision and active learning can indeed improve the predictive performance compared to weak supervision by itself.

\begin{figure}[h!]
    \centering
    \includegraphics[width=\textwidth]{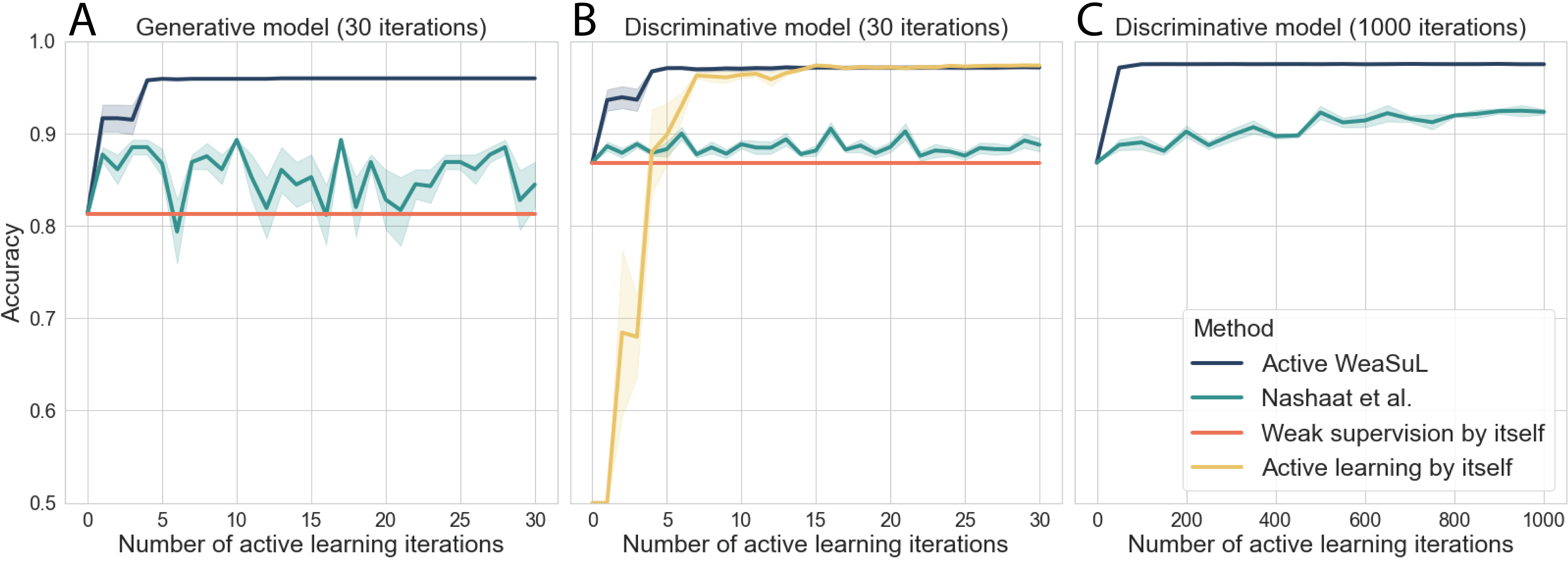}
    \caption{Predictive performance vs. active learning iterations on the artificial data set. Uncertainty bands were obtained by rerunning the  experiment 10 times with a different random seed, affecting the random selection of a data point within the bucket selected by maxKL. A\&B) Results for the first 30 active learning iterations for A) the generative models and B) discriminative models. C) Results for the discriminative models for Active WeaSuL and the method by \citet{nashaat2019hybridization} for 1000 active learning iterations.}
    \label{fig:artificial}
\end{figure}


Next, we compare Active WeaSuL to active learning by itself. To this end, we train a logistic regression model to predict $\textbf{y}$ from $\textbf{X}$, combined with an active learning sampling strategy based on distance from the decision boundary. Figure 3B shows how this approach compares to Active WeaSuL. While active learning eventually reaches the same performance as Active WeaSuL, it requires 9 additional active learning iterations to do so. We reason that Active WeaSuL may reach its optimal performance sooner due to having a warm start: the information from the weak labels allows it to start from an accuracy of 0.87 instead of an accuracy of 0.5. Altogether, this shows that, on these data, Active WeaSuL outperforms active learning by itself.

Finally, we compare Active WeaSuL to the competing method by \citet{nashaat2019hybridization}. In the first 30 active learning iterations, their approach achieves roughly the same accuracy as weak supervision, and hence does not outperform Active WeaSuL (Figure 3A\&B). When we extend the number of active learning iterations to 1000, their approach shows a slow increase in predictive performance, but still does not reach the performance that Active WeaSuL obtains after 4 iterations (Figure 3C). We note that in their own work, \citet{nashaat2019hybridization} only report performance after obtaining labels for thousands of data points, hence our observations here are consistent with the results they reported themselves. For an interpretation of why the approach by \citet{nashaat2019hybridization} requires so many more active learning iterations compared to Active WeaSuL, we refer to the Discussion section.

Overall, we have shown here that on the artificial data set illustrated in Figure 1, Active WeaSuL outperforms weak supervision and requires (far) fewer labelled data compared to either active learning or the approach by \citet{nashaat2019hybridization}.



\subsection{Application to realistic data sets}

We benchmark Active WeaSuL on two realistic data sets: a visual relationship detection task \citep{Lu2016} and a spam detection task \citep{Alberto2015}.

\begin{figure}[h!]
    \centering
    \includegraphics[width=\textwidth]{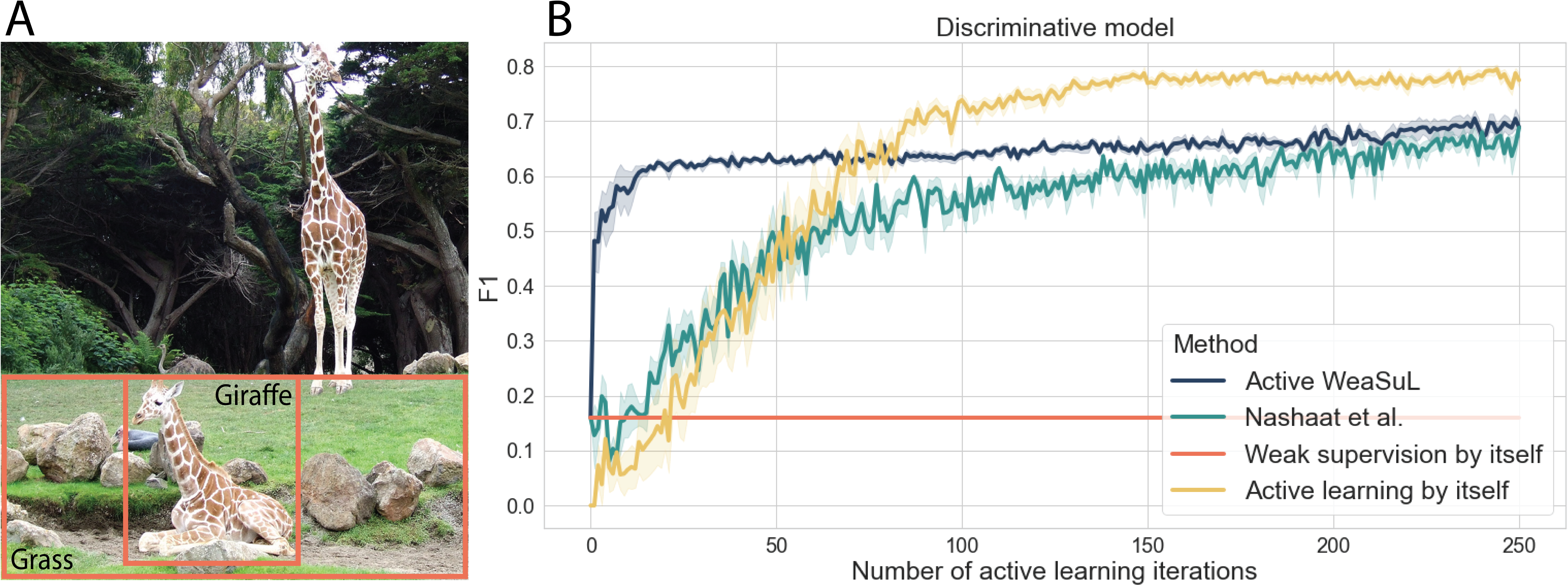}
    \caption{Predictive performance vs. active learning iterations on the visual relationship detection data set. A) Example of a visual relationship detection task: does the giraffe (subject) sit on the grass (object)? B) Results for the first 250 active learning iterations for the discriminative models.}
    \label{fig:vrd}
\end{figure}

In the visual relationship detection task, we predict whether a subject is sitting on top of an object based on: an image, bounding boxes of the subject and the object, and categorical variables indicating the type of object and subject (Figure \ref{fig:vrd}A). The data consist of 794 data points in the training set and 185 data points in the test set. We use the bounding boxes and categorical variables to define a set of three weak labels (Supplementary Materials). Here too, we set $\alpha=1$ in Equation \ref{eq:z-min-penalty}, as the loss of the initial generative model is close to zero.

Using these data, we compare the predictive performance of Active WeaSuL against the same set of methods as before. Because of the class imbalance, we use the F1 score to compare the methods. We focus on the predictive performance of the discriminative models, for which we use a neural network consisting of ResNet-18 \citep{He2016} for the image data and GloVe \citep{Pennington2014} for the categorical data (Supplementary Materials). 

Active WeaSuL sharply increases its performance from 0.16 to 0.62 within 16 iterations (Figure \ref{fig:vrd}B). Although active learning eventually trumps this performance, Active WeaSuL requires far fewer labelled data points to achieve a good predictive performance: it achieves a performance of 0.58 with only 7 iterations, whereas active learning requires 66 iterations and \citet{nashaat2019hybridization} requires 77 iterations to achieve the same performance. This shows that on these data, Active WeaSuL can obtain good (though not optimal) predictive performance using very few labelled data points, thereby making it a good choice in settings where obtaining labels for a large number of data points is difficult.

The spam detection task aims to classify whether YouTube comments are spam \citep{Alberto2015}. These data consist of 1,586 training samples and 250 test samples. The weak supervision model is based on a Snorkel tutorial\footnote{\url{https://github.com/snorkel-team/snorkel-tutorials/tree/master/spam}}, where we use the first 7 of the 10 labelling functions defined therein. Similarly, we train the discriminative model as defined in the tutorial.

For this task, we found that Active WeaSuL's initial generative model (i.e. before
any active learning iterations) had a relatively high loss of around 145. To make the penalty in Equation \ref{eq:z-min-penalty} competitive with the loss function, we suggest setting $\alpha = \max_{\textbf{z}_i} \mathrm{d}^2 L_G / \mathrm{d} \textbf{z}_i^2 $, where $L_G=||(\boldsymbol{\Sigma}^{-1} + \textbf{zz}^T)_{\Omega}||_F$ is the original loss of the generative model (left-hand term in Equation \ref{eq:z-min-penalty}) and $\textbf{z}_i$ are the fitted parameters in Equation \ref{eq:z-min-penalty}. This is statistically motivated by interpreting the original loss function as a chi-squared function that is minimized. This results in a penalty term (right-hand term of Equation \ref{eq:z-min-penalty}) that can compete with the original loss function. However, the best method to fine-tune $\alpha$ needs more study. Here we set the hyper-parameter $\alpha=10^6$.

\begin{figure}[h!]
	\centering
	\includegraphics[width=0.7\textwidth]{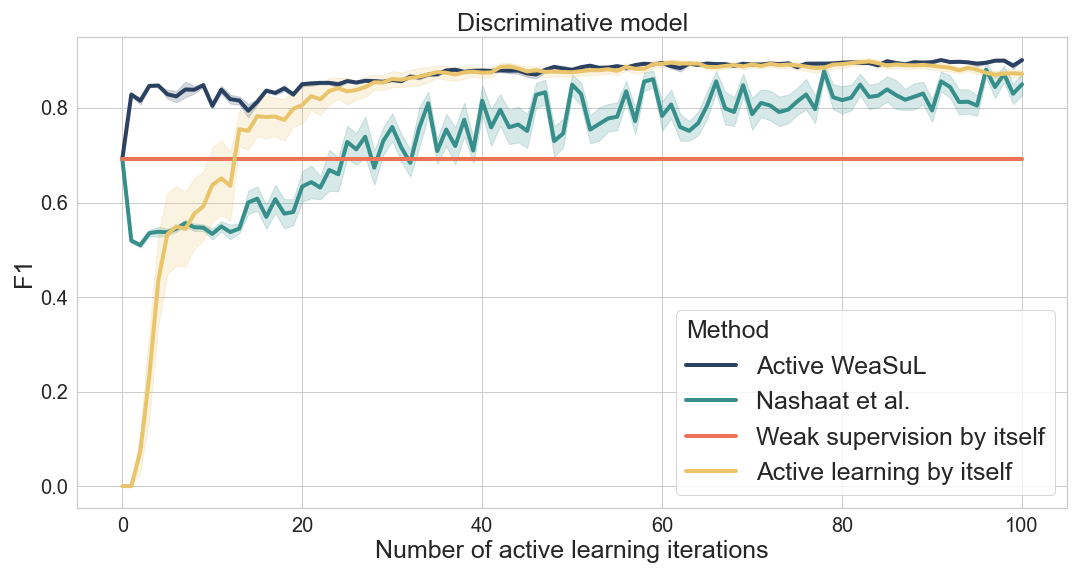}
	\caption{Predictive performance vs. active learning iterations on the spam detection data set. Results for the first 100 active learning iterations for the discriminative models.}
	\label{fig:spam}
\end{figure}

In Figure \ref{fig:spam} we compare the predictive performance of Active WeaSuL against the same set of methods as before.
We observe a similar pattern as for the other data sets: Active WeaSuL outperforms each of the baseline models. Its performance increases sharply from $F1=0.69$ to 0.85 within 4 iterations, which otherwise requires 27 iterations of active learning by itself, or 50 by \citet{nashaat2019hybridization}. We also observe that active learning does not overtake Active WeaSuL.

\subsection{Active learning sampling strategy}
Here, we zoom in on the benefits of our proposed active learning sampling strategy. To this end, we use the visual relationship detection data set to compare Active WeaSuL with three different sampling strategies: 1) maxKL divergence, our proposed strategy as described in Section \ref{sec:methods}; 2) margin, a strategy that samples data points whose probabilistic labels $\hat{p}_t(y | \boldsymbol{\lambda})$ are close to 0.5; and 3) random, a strategy that samples a data point from a random bucket.

Let us first consider the predictive performance of these various sampling strategies (Figure 6A). We find that the margin and random sampling strategies obtain the worst performance, suggesting that it is indeed beneficial to instead sample data points from buckets for which the model has trouble determining probabilistic labels.

\begin{figure}[t]
    \centering
    \includegraphics[width=\textwidth]{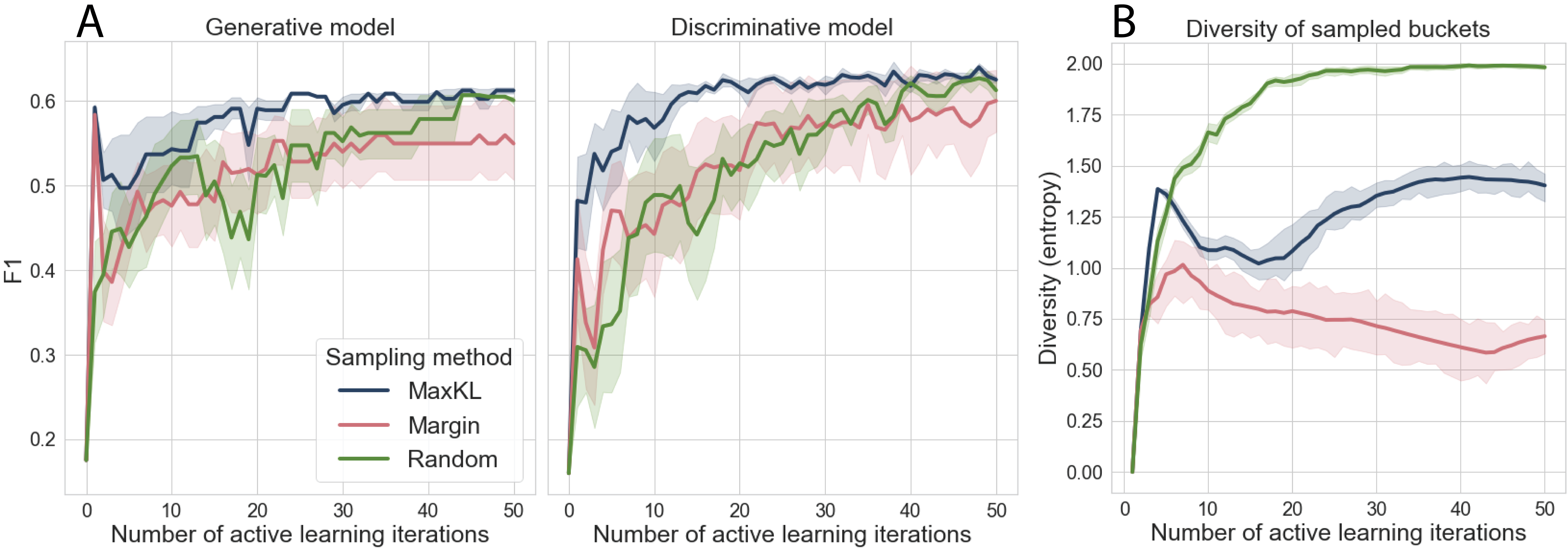}
    \caption{Comparison of the various sampling strategies on the visual relationship detection data set. A) The predictive performance for the generative model and the discriminative model. B) The diversity of the buckets that have been sampled.}
    \label{fig:sampling-strategies}
\end{figure}

An important shortcoming of the margin strategy is that it tends to repeatedly sample from the same bucket. This specifically happens when in a given bucket both classes are equally present, i.e. where 
$p_t(y | \boldsymbol{\lambda}) = 0.5$. Hence, compared to the other strategies, the margin strategy samples from a less diverse set of buckets, leading to labelled data that is less representative of the whole data set.

To quantify this behaviour, we define the following metric: given that we have sampled $t$ data points, and that for a given bucket $i$ we have sampled $n_i$ data points, consider $r_i = n_i \text{ / } t$ for each bucket $i$. We then summarise these fractions across buckets using entropy: $H = - \sum_i r_i \log r_i$. We note that entropy is high when the data points are sampled from a diverse set of buckets.

Indeed, we find that the margin strategy results in the lowest entropy, indicating that the data points were drawn from a non-diverse set of buckets (Figure 6B). Random sampling on the other hand leads to the highest entropy. This is in line with expectation, as random sampling should result in a highly representative set of data points (but not necessarily data points for which the model struggles). Our maxKL divergence approach results in entropy that lies between the two other approaches, suggesting that it balances the selection of representative data points with the selection of data points for which the model struggles.

\section{Discussion \& Conclusion}
For all three data sets, the combination of weak supervision and active learning works because the knowledge from the expert rules is clearly informative, providing a useful warm start to solve the supervised classification problem, which active learning then improves upon. In all examples, the defined weak labels give probabilistic labels that are valuable and decent prior estimates, as visible from the resulting performance measures, albeit (slightly) biased and thus still suboptimal. The penalty term, which incorporates the manually labelled data points, pulls the probabilistic labels closer to their correct values, resulting in improved classifier performance. We argue that this happens even with only a few labelled data points, as the correct solution may be nearby, giving flexibility to move the assigned probabilistic labels.

We have shown that Active WeaSuL outperforms active learning early on, for example with $\leq 60$ active learning iterations for the visual relationship detection data. We note however that active learning does eventually overtake Active WeaSuL. While early on the weak labels provide a warm start, we argue that in a later stage they may actually constrain the model too much: at some point a fully supervised model can simply learn more than a weakly supervised one. Hence, when it is possible to obtain a large number of labelled data points, one should use active learning. When the labelling budget is more limited, we suggest using Active WeaSuL.

All examples shown are straightforward classification problems where standalone active learning reaches optimal performance with only a limited number of manually labelled data points (13 for the artificial data; 150 and 40 for the realistic data). Interestingly, the number of manually labelled data points required for active learning to approach optimal performance on its own was higher in the (more complex) realistic data sets, suggesting that this number will further increase in even more complex problems. In such scenarios, the reduced labelling effort will also more prominently offset the initial effort of defining labelling functions. Altogether, we thus expect the added benefit of Active WeaSuL to further increase in more complex scenarios, specifically when the budget for requesting labelled data points is limited.

Finally, let us compare Active WeaSuL with the competing method by \citet{nashaat2019hybridization}. If for an expert-labelled data point $i$ we have $\boldsymbol{\lambda}_{i,*}=[1,1,0]$ and $\textbf{y}_i=1$, \citet{nashaat2019hybridization} correct the weak labels with the label provided by the expert, i.e. they assign $\boldsymbol{\lambda}_{i,*}=[1,1,1]$. While this clearly results in correct predictions for all expert-labelled data points, the effect on other data points is small and not necessarily beneficial. For example, the above correction increases the overall amount of agreement between weak labels, resulting in a slightly higher contribution of the third weak label $\boldsymbol{\lambda}_{*,3}$ across data points, even though it was actually mistaken here. In Active WeaSuL on the other hand, the penalty introduced in Equation \ref{eq:z-min-penalty} will actually decrease the overall contribution of $\boldsymbol{\lambda}_{*,3}$, exactly because it was mistaken. Hence, we argue that the approach of \citet{nashaat2019hybridization} requires a large amount of active learning iterations because it relies on corrections that mostly affect individual data points, whereas Active WeaSuL requires fewer active learning iterations because it uses the expert-labelled data to learn how to best combine the weak labels.

Altogether, Active WeaSuL provides a means to make efficient use of domain expert knowledge: by combining expert-defined rules (weak supervision) with small amounts of expert-labelled data (active learning), we can train supervised models in applications where obtaining labelled data is inherently difficult.



\subsubsection*{Acknowledgements}
We would like to thank the ING Wholesale Banking Advanced Analytics group, and in particular Ilan Fridman Rojas and Fabian Jansen, for their support and constructive feedback during this project.

\bibliography{bibliography}
\bibliographystyle{iclr2021_conference}

\appendix
\newpage
\section{Appendix}


\subsection{Active WeaSuL algorithm}

\begin{algorithm}[h!]
    \label{algo:algorithm}
    \SetAlgoLined
    \textbf{Input:} Unlabelled data set $\textbf{X}$, active learning budget $B$ \\
    Create labelling functions and apply them to the $\textbf{X}$ to obtain the weak labels $\boldsymbol{\lambda}$\;
    Fit generative model (Equation \ref{eq:z-min}) to get the initial parameters $\hat{\textbf{z}}$\;
    Compute the probabilistic labels $\hat{p}_{t=0}(y|\boldsymbol{\lambda}) = f(\hat{\textbf{z}}, \boldsymbol{\lambda})$\;
    \For{$t$ $\leftarrow$ 1 to $B$}{
        Using Equations \ref{eq:prob-labels-t}-\ref{eq:kl-div}, determine the divergence $\text{KL}_{t,i}$ for all buckets $i$\;
        Sample a data point from the bucket whose difference $\text{KL}_{t,i}$ is the largest (maxKL)\;
        Have an expert label the sampled data point, add it to the set of expert-labelled data $D$\;
        Fit generative model again, now using Equation \ref{eq:z-min-penalty}, resulting in an updated $\hat{\textbf{z}}$\;
        Compute $\hat{p}_{t}(y|\boldsymbol{\lambda}) = f(\hat{\textbf{z}}, \boldsymbol{\lambda})$\;
        Optionally, fit a discriminative model using feature matrix $\textbf{X}$ and target $\hat{p}_{t}(y|\boldsymbol{\lambda})$
    }
    \textbf{Output:} Probabilistic labels $\hat{p}_{t}(y|\boldsymbol{\lambda})$
    \caption{Active WeaSuL}
\end{algorithm}

\subsection{Discriminative model}
It is often possible to further improve on the generative model's predictive performance by training a discriminative model that predicts the probabilistic labels $\hat{p}_t(y | \boldsymbol{\lambda})$ from the feature matrix $\mathbf{X}$. This essentially allows one to define a more granular decision boundary (Figure 1D). To train the discriminative model on probabilistic labels, we use the formulation from \citet{ratner2017snorkel}.

We incorporate the information from active learning into the discriminative model in two ways. First, at each active learning iteration $t$, we train the discriminative model using the probabilistic labels from the generative model at iteration $t$. Second, for all expert-labelled data points in $D$, we use the binary label provided by the expert rather than the probabilistic label. Note that we do not use the discriminative model to determine which data points should be labelled next.

\subsection{Visual Relationship Detection}
We have based our Visual Relationship Detection (VRD) experiment on one of the tutorials\footnote{\url{https://github.com/snorkel-team/snorkel-tutorials/tree/master/visual_relation}} in Snorkel, a popular Python package for weak supervision \citep{ratner2017snorkel}. To perform this task with Active WeaSuL, we have made a few changes, which we describe here.

While the original VRD task is a multi-class experiment, where each class represents a relationship between a subject and an object (e.g. \textit{sitting on, on top of, taller than}), we focus here on only one class in particular: is the subject sitting on top of an object? This turns the problem into a binary classification problem, which we can then tackle with Active WeaSuL. Since the VRD data contain quite a few duplicates (i.e. the same data point with different labels), we have removed all duplicates. We have then labelled a data point as $y=1$ when one of its duplicates had the label \textit{sitting on} and $y=0$ otherwise. Finally, we have changed the labelling functions such that they are tailored towards the binary classification task (Supplementary Table 1). Note that the first two weak labels are conditionally dependent (as they are based on the same underlying data), and we incorporate this information in the generative model via $\Omega$ (Equation \ref{eq:z-min}).
\begin{table}[h!]
    \centering
    \caption{Labelling functions for classifying \textit{sitting on} for the VRD task.}
    \begin{tabularx}{\textwidth}{ m{0cm} m{5.2cm} X }
        \hline
         & \textbf{Description} & \textbf{Labelling function} \\
         \hline
        1 & relative bounding box sizes & \texttt{subject.area / object.area < 0.8} \\
        2 & distance between subject and object & \texttt{dist(subject, object) > 100} \\
        3 & object and subject category & \texttt{subject.type == 'person' AND \newline object.type IS IN ['bench', 'chair', 'floor', 'horse', 'grass', 'table']} \\
        \hline
    \end{tabularx}
    \label{tab:vrdpatterns}
\end{table}

As in the Snorkel tutorial, we have used a neural network for our discriminative model (consisting of ResNet-18 \citep{He2016} for the image data and GloVe \citep{Pennington2014} for the categorical data).

To prevent overfitting, we split the training set such that 10\% of the probabilistic labels are kept aside as a validation set, to be used for early stopping. We then train the model for a maximum of 100 epochs per active learning iteration, from which we choose the parameter setting corresponding to the minimal loss. The training is stopped when the minimal loss does not improve in five consecutive epochs, i.e. $patience=5$. Note that we do not leak information in this way: we only use probabilistic labels to create the validation set, not the ground truth labels. Finally, note that the validation set is separate from the test set, hence our estimate of the predictive performance remains unbiased.

\end{document}